\documentclass[journal]{IEEEtran}
\usepackage{amsmath,amsfonts}
\usepackage{algorithmic}
\usepackage{algorithm}
\usepackage{array}
\usepackage[caption=false,font=normalsize,labelfont=sf,textfont=sf]{subfig}
\usepackage{textcomp}
\usepackage{stfloats}
\usepackage{url}
\usepackage{verbatim}
\usepackage{graphicx}
\usepackage{cite}
\usepackage{color,soul}
\usepackage{enumitem}
\usepackage{booktabs}
\begin{document}

\title{A Cooperative Aerial System of A Payload Drone Equipped with Dexterous Rappelling End Droid for Cluttered Space Pickup}

\author{Wenjing Ren,~Xin Dong*,~Yangjie Cui,~Binqi Yang,~Haoze Li,~Tao Yu,\\~Jinwu Xiang,~Daochun Li,~and Zhan Tu*
\thanks{Wenjing Ren,~Yangjie Cui,~Binqi Yang,~Haoze Li,~Tao Yu,~Jinwu Xiang,~and Daochun Li are with the School of Aerospace Science and Engineering,~Beihang University,~Beijing 100191,~China.}
\thanks{Xin Dong is with the Hangzhou International Innovation Institute of Beihang University,~Hangzhou 311115,~China.~\texttt{Email: xindong324@buaa.edu.cn}}
\thanks{Zhan Tu is with the Institute of Unmanned System,~Beihang University,~Beijing 100191,~China.~\texttt{Email: zhantu@buaa.edu.cn}}
\thanks{Jingwu Xiang, Daochun Li, and Zhan Tu are also with the Tianmushan Laboratory, Yuhang District,~Hangzhou 311115,~China.}
\thanks{* Author for correspondence.}}

\markboth{ }%
{Shell \MakeLowercase{\textit{et al.}}: A Sample Article Using IEEEtran.cls for IEEE Journals}


\maketitle

\begin{abstract}
In cluttered spaces, such as forests, drone picking up a payload via an abseil claw is an open challenge, as the cable is likely tangled and blocked by the branches and obstacles. To address such a challenge, in this work, a cooperative aerial system is proposed, which consists of a payload drone and a dexterous rappelling end droid. The two ends are linked via a Kevlar tether cable. The end droid is actuated by four propellers, which enable mid-air dexterous adjustment of clawing angle and guidance of cable movement. To avoid tanglement and rappelling obstacles, a trajectory optimization method that integrates cable length constraints and dynamic feasibility is developed, which guarantees safe pick-up. A tether cable dynamic model is established to evaluate real-time cable status, considering both taut and sagging conditions. Simulation and real-world experiments are conducted to demonstrate that the proposed system is capable of picking up payload in cluttered spaces. As a result, the end droid can reach the target point successfully under cable constraints and achieve passive retrieval during the lifting phase without propulsion, which enables effective and efficient aerial manipulation.
\end{abstract}
\begin{IEEEkeywords}
Cooperative aerial system, Drone pickup, Rappelling end droid, Cluttered space
\end{IEEEkeywords}

\section{Introduction}
\IEEEPARstart{W}{ith} the rapid development of unmanned aerial vehicles (UAVs), UAV-based aerial transportation has garnered increasing attention. 
Two primary payload picking up strategies are commonly employed, i.e., direct attachment to the UAV and abseiling picking up\cite{ref2}. 
For the direct attachment approach, the payload needs to be mounted beneath the UAV using grippers or robotic arm mechanisms. 
Such a design requires the UAV to land for cargo loading, the effectiveness of which is often limited by the landing surroundings\cite{ref19}.

In contrast, abseiling-based package picking can work at much higher altitude and does not require UAV landing. Regardless, it also requires an open environment and a clear view of the target payload\cite{ref1}. In cluttered spaces, such as forests, drone picking up a payload via an abseil claw is an open challenge, as the cable is likely tangled and blocked by the branches and obstacles. Planning an efficient and safe rappelling trajectory is essential. In addition, the end of the grappling must be controlled to accurately follow the trajectory to capture the cargo.

\begin{figure}[!t]
    \centering
    \includegraphics[width=3.6in]{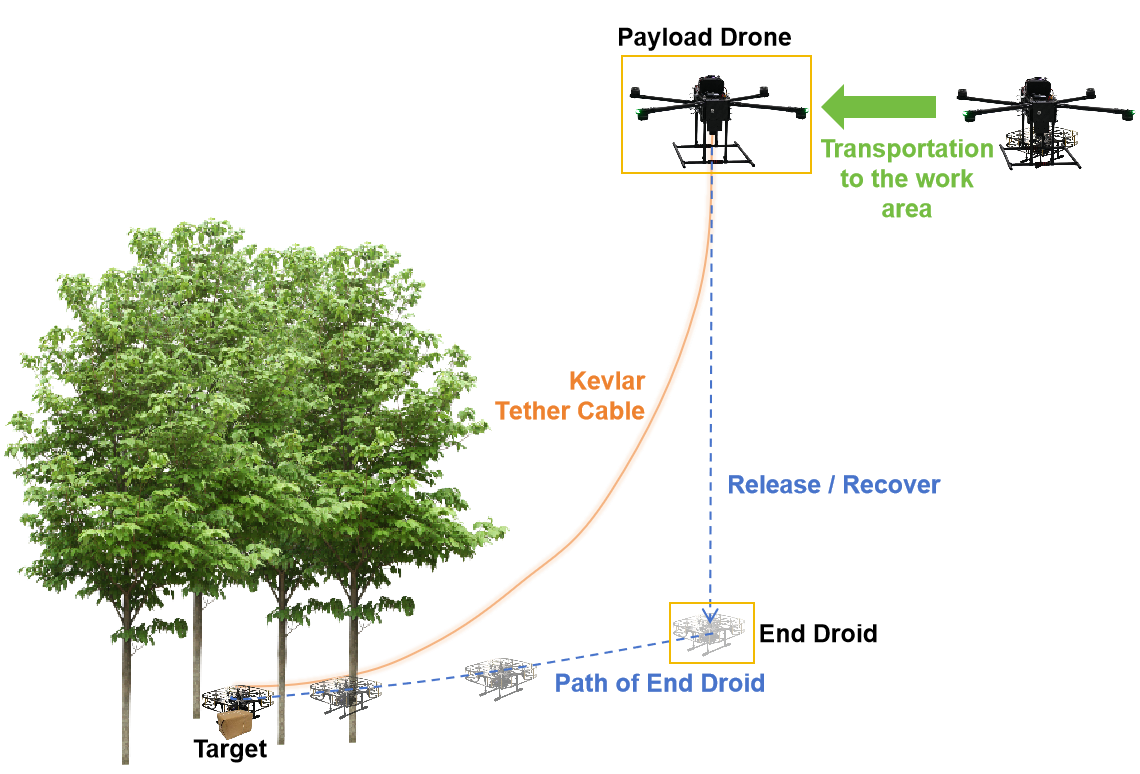}
    \caption{Overall architecture and typical pick-up scenario of the proposed cooperative aerial system. The system consists of a payload drone and a dexterous rappelling end droid. In cluttered spaces such as forests, a payload drone with sufficient carrying capacity cannot enter to pick up the target. The dexterous rappelling end droid could freely pass through narrow spaces and locate the target.}
    \label{fig: working_process}
\end{figure}

In this paper, considering the target picking problem under cluttered environments, we present a cooperative aerial system equipped with a dexterous rappelling end droid. To minimize the impact of the disturbance from the cable, the cable should neither be too tight nor too slack. Based on such respects, a catenary model is introduced to establish the maximum safe cable length constraint. The catenary model accurately characterizes the natural sagging shape of a flexible cable under its own weight. Thus, it is reasonable to evaluate the cable's configuration under various relative positions. This helps ensure that the tether remains within a safe range throughout motion, avoiding issues such as slackness, oscillation, or entanglement. Therefore, the motion trajectory of the end droid is planned by constraining it in a certain solution space to ensure flight safety. Through spatial-temporal joint trajectory optimization, the end droid achieves rapid and agile landing on the target. Experimental studies have been conducted to validate the effectiveness of the proposed system. As a result, the payload drone cooperates with the end droid properly. Such a design demonstrates a complete operation pipeline, including cable-guided descent, constrained trajectory planning and tracking, target pick-up, and retrieval. 

The main contributions of this paper are as follows:

\begin{enumerate}[label=\arabic*), left=0pt, align=left, itemsep=0.5ex]
  \item A cable-suspended aerial system is proposed, which consists of a payload drone, an end droid, and a connecting cable, as illustrated in Fig.~\ref{fig: working_process}. Compared with traditional UAV-mounted suspension mechanisms, the proposed system offers an expanded operational workspace. Compared to conventional transport-end droid configurations, it also avoids the complexity of mid-air retrieval and the adverse aerodynamic effects caused by the downwash of the transport UAV.

  \item The cable is modeled using a catenary formulation, which more accurately captures the natural sagging behavior of the cable and accounts for its influence on end droid motion.

  \item A trajectory optimization method that enables efficient motion planning under cable constraints is proposed. The method is validated both in simulation and real-world experiments, demonstrating its reliability under quasi-static and dynamic conditions.

\end{enumerate}

The rest of this paper is organized as follows. Section II reviews related research on UAV systems equipped with end-effectors. Section III provides a detailed description of the cable modeling process and develops a process model for the controlled release of the end droid under length constraints at a constant descent speed. Section IV formulates the cable length constraint as a nonlinear optimization problem and proposes a corresponding real-time optimization strategy. Section V presents and analyzes the results from both simulation and experimental evaluations. Finally, Section VI concludes the study and outlines potential directions for future research.

\section{Related works}
In the field of object picking, two primary methods have been extensively explored within the existing research, namely, aerial manipulator UAV and cable-suspended UAV. The aerial manipulator UAV is known for its agility and precision in picking operations, while the cable-suspended UAV offers distinct advantages in long-distance picking tasks. 

\subsection{Aerial Manipulator UAV}
Aerial manipulator UAVs, equipped with robotic arms, have been developed to perform a variety of tasks such as grasping, manipulation, and inspection. 
Research in this domain has explored diverse design approaches. Underside and topside manipulator mounting positions have been investigated to address specific inspection requirements\cite{ref5,ref7}. An upward-facing design tailored for torsion operations has been introduced\cite{ref8}. Meanwhile, control strategies for stable physical interaction have been established, laying a groundwork for precise aerial manipulation\cite{ref6}. These foundational efforts have enabled more specialized advancements. A tightly integrated system for valve-turning tasks has been developed\cite{ref9}. An innovative continuum manipulator, the Aerial Elephant Trunk, designed for versatility in constrained environments, has also been proposed\cite{ref22}.
UAVs equipped with manipulators could achieve dexterous picking in the mid-air, while its operating space is limited by the size of the manipulators, and it cannot perform long-distance operations at high altitudes. Furthermore, the mechanical arm that is directly affixed to the fuselage of the UAV can compromise flight stability during missions.

\subsection{Cable-Suspended UAV}
Wire-suspended UAVs leverage tethers or cables to expand operational capabilities across diverse applications. 
A multirotor UAV featuring a suspension device for high-altitude pruning is proposed\cite{ref14}. 
A multimodal airborne tethered robot has been designed\cite{ref21}, which combines the capabilities of aerial flight and tethered movement, and is capable of autonomous navigation and bypassing obstacles in the forest canopy layer, used to obtain environmental data in hard-to-reach areas.
Currently, the end of cable-suspended UAVs are mainly categorized into electromagnetic, gripper, and shooting types.
A grasping mechanism based on electromagnets installed under UAVs is designed to enhance the interaction ability between UAVs and the environment, especially in scenarios where rapid and reliable grasping is required\cite{cable-mag-uav-2020}. But its drawback lies in that it can only grasp magnetic objects. 
For the precise adoption at the end, a fan-based gripper form at the end has been proposed\cite{cable-ductedfan-uav-2019, cable-ductedfan-uav-2020}, which uses two wires and four ducted fans with an imu to suppress the oscillation at the end. 
A device combining UAV and shooting operating arms is proposed\cite{ref17}, which is capable of efficiently grasping target objects at long distances and in complex environments with obstacles through pulse force transmitters, recovery devices and end effectors.
However, these ends lack of active motion capabilities, which limits their ability to perform long-distance picking in obstacle-dense environments.

The aerial manipulator UAV is capable of agile and precise grasping operations, making it suitable for tasks that require direct interaction with objects in close proximity.  However, it faces limitations in long-distance operations.  On the other hand, the cable-suspended UAV has the advantage of being able to perform long-distance picking tasks and exhibits greater stability in complex environments.  Nevertheless, its powerless end-effector restricts picking operations in obstacle-dense environments. To combines the strengths of both methods while addressing their respective limitations, a novel cooperative aerial system is proposed, which is equipped with a dexterous rappelling end droid and adjustable cable length to achieve dexterous and long-distance pickup in cluttered space.

\section{Problem Statement and Framework}
This section introduces the overall system architecture, including the model of the payload drone, the model of cable, and the pickup process in cluttered space.
\subsection{Framework}
\begin{figure}[t!]
\centering
\includegraphics[width=\linewidth]{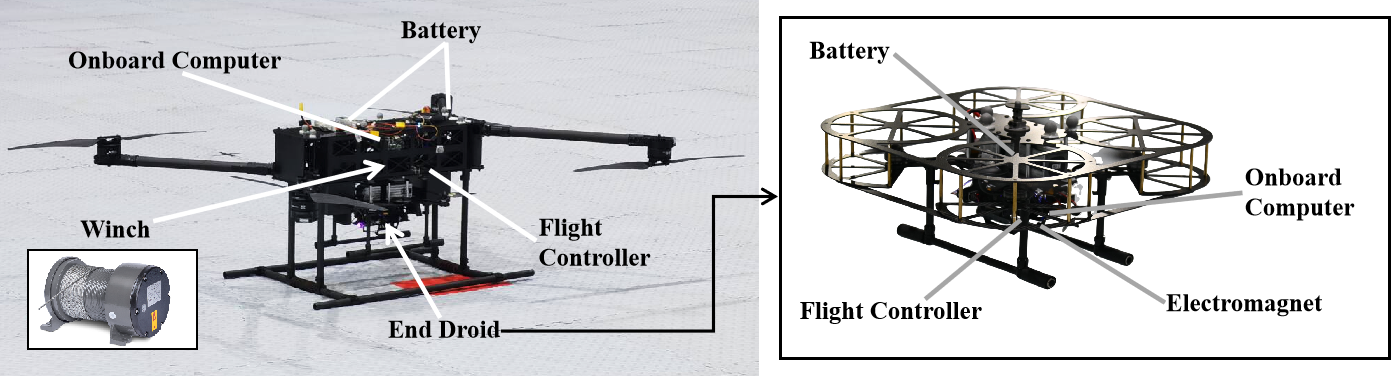}
\caption{Platform of the payload drone equipped with a dexterous rappelling end droid.}
\label{fig: platform}
\end{figure}

The overall framework of the hardware platform used in this paper is shown in Fig.~\ref{fig: platform}. The system consists of a payload drone and a dexterous rappelling end droid, connected via a Kevlar tether cable. The payload drone is a quadrotor equipped with a winch mounted beneath it, which controls the retraction and extension of the cable. The end droid is actuated by four upward-facing propellers, similar to a quadrotor, enabling it to maneuver flexibly in cluttered environments and to guide the cable during the descent. Its wheelbase is 380~mm. A model-based nonlinear controller is designed to stabilize the payload drone and ensure accurate trajectory tracking simontaneously. Both drones are tracked in real-time by an OptiTrack motion capture system, providing high-precision 3D localization. All trajectory planning and control computations are executed onboard using an RK3588 processor. 

The picking up process is shown in Fig.~\ref{fig: working_process}. After the cooperative aerial system fly to the working area, the winch of payload drone gradually releases the cable to make the end droid descend. The end droid begins to move autonomously through the cluttered space and guides the cable to move. Once the target was detected, the droid picked up the target object and return. Finally, the winch retrieves the cable to retrieve the end droid and the target object.

Since the cable gets entangled with obstacles during the process of the end droid traverse the cluttered space, this paper focus on the trajectory optimization of the end droid constrained by the cable. 

\subsection{Dynamic model of end droid}
The payload drone is modeled as a quadrotor UAV equipped with a winch. Its configuration is defined by its position $\boldsymbol{p}_d \in \mathbb{R}^3$, rotation matrix $\boldsymbol{R}_d \in SO(3)$, and control inputs including total thrust $f_d$ and body angular velocity $\boldsymbol{\omega}_d \in \mathbb{R}^3$. Assuming quasi-static winch behavior and neglecting winch dynamics, the motion equations are given by:

\begin{equation}
\begin{cases}
\boldsymbol{\tau}_d = \frac{f_d \boldsymbol{R}_d}{m_d}, \\
\ddot{\boldsymbol{p}}_d = \boldsymbol{\tau}_d - \boldsymbol{g} + \frac{1}{m_d} \boldsymbol{F}_{\text{tether}}, \\
\dot{\boldsymbol{R}}_d = \boldsymbol{R}_d \hat{\boldsymbol{\omega}}_d,
\end{cases}
\end{equation}
where $m_d$ is the mass of the payload drone, $\boldsymbol{g}$ is the gravitational acceleration vector, $\boldsymbol{F}_{\text{tether}}$ is the tension force from the tether cable, and $\hat{\boldsymbol{\omega}}_d$ denotes the skew-symmetric matrix of $\boldsymbol{\omega}_d$.

As the end droid has a quadrotor-like structure, its dynamic model follows the standard quadrotor dynamics. Let $\boldsymbol{p}_e \in \mathbb{R}^3$ denote its position, and $\boldsymbol{R}_e \in SO(3)$ the rotation matrix. With input of thrust $f_e$ and angular velocity $\boldsymbol{\omega}_e \in \mathbb{R}^3$, its motion equations are:

\begin{equation}
\begin{cases}
\boldsymbol{\tau}_e = \frac{f_e \boldsymbol{R}_e}{m_e}, \\
\ddot{\boldsymbol{p}}_e = \boldsymbol{\tau}_e - \boldsymbol{g} - \frac{1}{m_e} \boldsymbol{F}_{\text{tether}}, \\
\dot{\boldsymbol{R}}_e = \boldsymbol{R}_e \hat{\boldsymbol{\omega}}_e,
\end{cases}
\end{equation}
where $m_e$ is the mass of the end droid, and $-\boldsymbol{F}_{\text{tether}}$ is the reaction force exerted by the tether.

Due to its quadrotor-like structure, the end droid exhibits differential flatness properties\cite{mellinger2011minimum, faessler2018differential}. This means that all system states and control inputs can be represented using a set of flat outputs and their finite-order derivatives. The flat outputs of the end droid are chosen as:

\begin{equation}
\boldsymbol{y}_e =
\begin{bmatrix}
x_e \\
y_e \\
z_e \\
\psi_e
\end{bmatrix},
\end{equation}
where $\psi_e$ denotes the yaw angle. This property allows the end droid’s trajectory to be planned in the space of flat outputs, enabling smooth polynomial parameterization while satisfying dynamic feasibility and facilitating the inclusion of cable length and obstacle avoidance constraints in the optimization framework.

\subsection{Math model of cable}
\begin{figure}[!t]
\centering
\includegraphics[width=7cm]{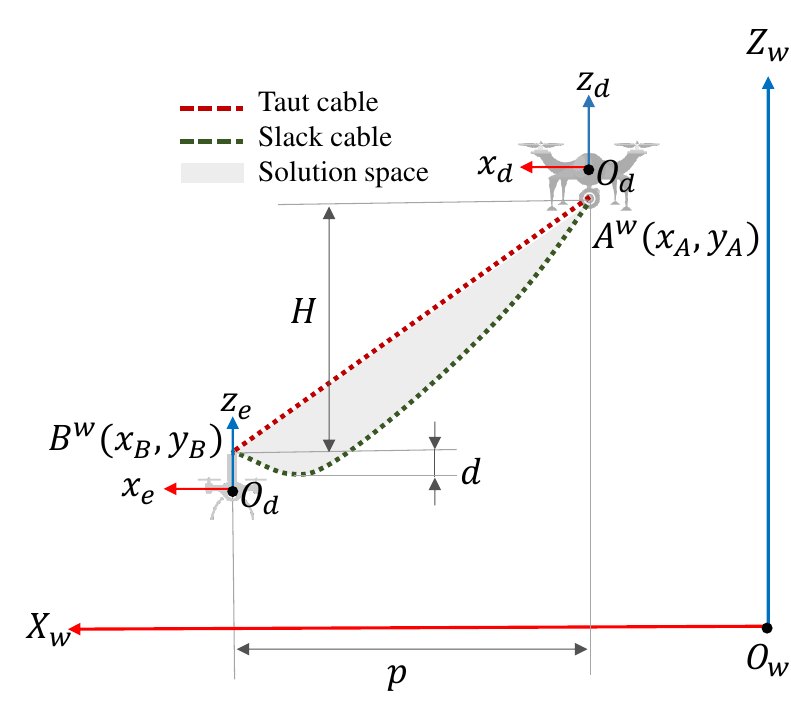}
\caption{Coordinate system setup and catenary-based modeling of the tether cable between the payload drone and the end droid.}
\label{fig: math_model_cable}
\end{figure}
Most existing studies assume that the cable remains continuously taut, focusing primarily on the attitude and trajectory control of the end while paying little attention to the detailed modeling of the cable itself. The tension in the cable is often treated as an external disturbance acting on the drone system, without adequately considering the variations in the shape and tension of the cable. Inspired by\cite{ref25}, a flexible cable is modeled in this study to connect the end droid and the payload drone, where the cable may alternate between sagging and taut states during system operation. To model the natural sag of the cable under its own weight, a catenary model is adopted. The two endpoints of the catenary are located above the end droid and below the payload drone, respectively. To avoid interference between the cable and the end droid’s rotors, a vertical protrusion of height $d$ is added to the top of the end droid. The endpoints of the catenary exhibit a certain horizontal and vertical distance. As shown in Fig.~\ref{fig: math_model_cable} , an East-North-Up (ENU) world coordinate system ${O_w}$ is established, with the $X_w$ and $Z_w$ axes representing the horizontal and vertical directions, respectively. To simplify the problem, we assume that both the payload drone and the end droid move only within the $X$–$Z$ plane, neglecting motion along the $Y$ direction. A local body-fixed coordinate frame ${O_d, x_d, z_d}$ is attached to the payload drone, and another local frame ${O_e, x_e, z_e}$ is attached to the end droid. These frames are used to describe the relative positions and orientations of the two platforms, as well as the configuration of the tether cable in the 2D plane. Let the coordinates of the end droid be denoted by $A^w(x_A, z_A)$, the initial position of the end droid (also the payload drone’s position) by $B^w(x_B, z_B)$, and the catenary’s vertex coordinates by $O(x_0, z_0)$, which vary as the relative positions of the cooperative aerial system change. For points A and B, the following catenary equations hold:
\begin{equation}\label{eq.2}
z_A=\frac{T_0}{\mu}\left(\cosh \frac{\mu x_A}{T_0}-1\right),
\end{equation}
\begin{equation}\label{eq.3}
z_B=\frac{T_0}{\mu}\left(\cosh \frac{\mu x_B}{T_0}-1\right),
\end{equation}

Here, $T_0$ is the tension at the vertex of the catenary, and $\mu$ is the weight per unit length of the cable. The above equations are applicable to any point between points $A$ and $B$. Additionally, points $A$ and $B$ satisfy the following relationships:
\begin{equation}\label{eq.4}
x_B=x_A+p,
\end{equation}
\begin{equation}\label{eq.5}
z_B=z_A+H.
\end{equation}

From  (\ref{eq.2}) to (\ref{eq.5}), it follows that:
\begin{equation}\label{eq.6}
H=\frac{T_0}{\mu}\left(\cosh \frac{\mu x_B}{T_0}-\cosh \frac{\mu\left(x_B-p\right)}{T_0}\right).
\end{equation}

On the other hand, the derivative of the angle equation is the slope equation of the curve. Applying the slope equation at points $A$ and $B$ yields:
\begin{equation}\label{eq.7}
\tan \theta_A=\left.\frac{d y}{d x}\right|_{x=x_A}=\sinh \frac{\mu x_A}{T_0},
\end{equation}
\begin{equation}\label{eq.8}
\tan \theta_B=\left.\frac{d y}{d x}\right|_{x=x_B}=\sinh \frac{\mu x_B}{T_0}.
\end{equation}

Moreover, the tension at any point can be expressed in terms of its tangential angle:
\begin{equation}\label{eq.9}
T=\frac{T_0}{\cos \theta},
\end{equation}
that is, for any point along the cable, the longitudinal tension is related to the cable angle through this equation.
\begin{figure}[!t]
\centering
\subfloat[]{\includegraphics[width=1.7in]{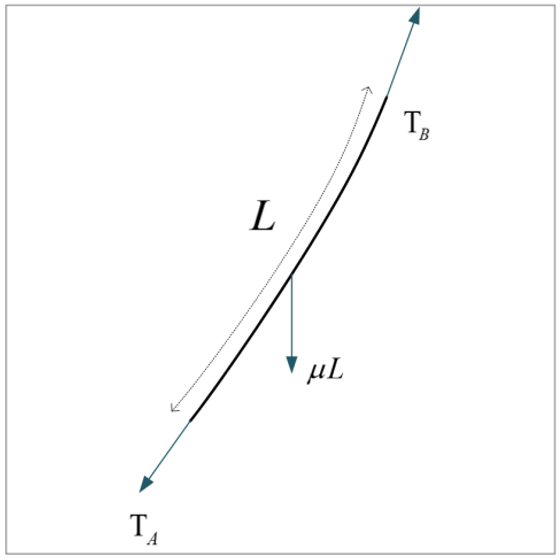}%
\label{without_sag}}
\subfloat[]{\includegraphics[width=1.7in]{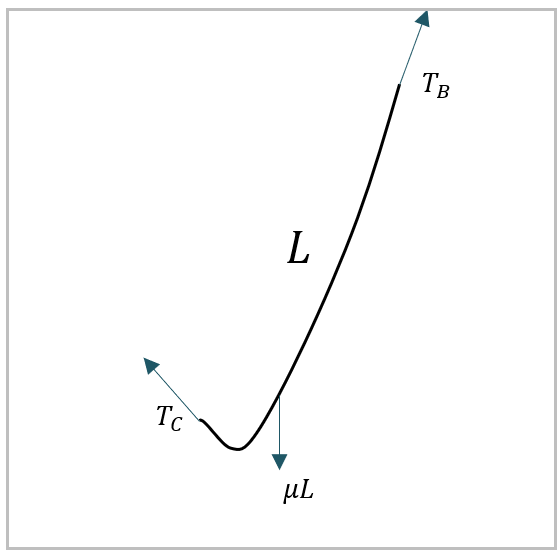}%
\label{with_sag}}
\caption{Force analysis diagram of the entire cable in two equilibrium states. (a) Taut cable. (b) Slack cable.}
\label{fig_cable_force}
\end{figure}

As shown in the Fig.~\ref{fig_cable_force}, these are the force analysis diagrams of the cable under two equilibrium states. The difference between Fig.~\ref{fig_cable_force}(a) and Fig.~\ref{fig_cable_force}(b) lies in whether the cable has sag. For the case where the cable has no sag, the equilibrium equations in the $x$ and $z$ directions can be established as follows:
\begin{equation}\label{eq.10}
\left\{\begin{array}{l}
T_A \cos \theta_A=T_B \cos \theta_B=T_0, \\
T_B \sin \theta_B=T_A \sin \theta_A+\mu L.
\end{array}\right.
\end{equation}

It can be obtained that:
\begin{equation}\label{eq.11}
\tan \theta_B=\tan \theta_A+\frac{\mu L}{T_0},
\end{equation}

Substituting into the  (\ref{eq.7}) to (\ref{eq.8}) yields:
\begin{equation}\label{eq.12}
L=\frac{T_0}{\mu}\left\{\sinh \frac{\mu x_B}{T_0}-\sinh \frac{\mu\left(x_B-p\right)}{T_0}\right\}.
\end{equation}

For the case where the cable has a certain amount of sag, the equilibrium equations become:
\begin{equation}\label{eq.13}
\left\{\begin{array}{l}
T_A \cos \theta_A=T_B \cos \theta_B=T_0, \\
T_B \sin \theta_B+T_A \sin \theta_A=\mu L.
\end{array}\right.
\end{equation}

Substituting into the  (\ref{eq.7}) to (\ref{eq.8}) yields:
\begin{equation}\label{eq.14}
L=\frac{T_0}{\mu}\left\{\sinh \frac{\mu x_B}{T_0}+\sinh \frac{\mu\left(x_B-p\right)}{T_0}\right\}.
\end{equation}

The above  (\ref{eq.6}), (\ref{eq.13}) and (\ref{eq.14}) can be combined to form the following system of equations:
\begin{equation}
\left\{\begin{aligned}
& H=\frac{T_0}{\mu}\left(\cosh \frac{\mu x_B}{T_0}-\cosh \frac{\mu\left(x_B-p\right)}{T_0}\right), \\
& L=\frac{T_0}{\mu}\left(\sinh \frac{\mu x_B}{T_0}-\sinh \frac{\mu\left(x_B-p\right)}{T_0}\right), \text { taut cable }, \\
& L=\frac{T_0}{\mu}\left(\sinh \frac{\mu x_B}{T_0}+\sinh \frac{\mu\left(x_B-p\right)}{T_0}\right), \quad \text { slack cable }.
\end{aligned}\right.
\end{equation}

To determine the solution of the system, these two equations need to be solved. The system contains five variables: $L$, $p$, $H$, $x_B$, and $T_0$, where $p$, $H$, and $x_B$ are known, while $L$ and $T_0$ are the unknowns to be solved.

\subsection{Modeling of end droid operation process}
In cable-constrained end droid motion, proper control of the cable length is critical to ensuring system safety and stability. If the cable is too short, it may exert undesired pulling forces on the payload drone, negatively impacting its attitude and trajectory control. Conversely, if the cable is too long, excessive sagging may result in drag or entanglement during flight, which can restrict the maneuverability of the end droid and introduce safety risks.

To address this, a cable modeling approach based on the catenary curve is adopted in this study. By using the relative position between the end droid and the payload drone, the model enables real-time calculation of the maximum and minimum allowable cable lengths that prevent both excessive tension and rotor interference. This defines a dynamic constraint interval for cable length throughout the end droid’s motion. The end droid need to maintain the cable length within this range during its flight to ensure safe and controlled operation, ultimately reaching the target point.

The reachable region of the end droid under cable constraints is determined by this dynamic interval, as illustrated in Fig.~\ref{fig: math_model_cable}, and it defines the feasible domain for trajectory planning.

\section{Trajectory Optimization Constrained by Cable Length}
\subsection{Trajectory representation}
Based on this modeling and constraint framework, this paper proposes a trajectory planning method for the end droid. The method begins by generating an initial, collision-free path from the current position to the target, taking into account the target’s position and motion intention. This path is then refined and adjusted into a dynamically feasible trajectory that satisfies both tension and cable length constraints throughout the flight, ensuring physical plausibility and stable execution.

After obtaining the initial trajectory, the trajectory is optimized by adopting the MINCO\cite{ref18}, a minimum control effort polynomial trajectory class. The trajectory can be obtained through a linear-complexity map from the intermediate points $q$ and the duration vector $\boldsymbol{T}$ to the coefficients of the trajectory $c=\mathcal{M}(q, \boldsymbol{T})$. In this paper, an N-segment uniform MINCO trajectory is used, which means the duration of each segment $dT$ is equal, the total duration of whole trajectory is $\boldsymbol{T}=N \cdot d T$. The advantage of the MINCO trajectory is that given the cost function $\boldsymbol{J}(q, T)=F(q, T)$, MINCO gives an efficient way to obtain the gradient $\partial \mathcal{J} / \partial q$  and $\partial \mathcal{J} / \partial T$, enabling efficient trajectory optimization given the penalty cost function.

For the UAV motion under cable constraint problem, the initial MINCO trajectory is constructed by sampling N intermediate points from the initial trajectory. Subsequently, the entire trajectory is optimized based on the constraint conditions.

\subsection{Trajectory optimization}

Inspired by\cite{ref24}, for the cable UAV trajectory optimization problem, given the motion of the end droid $p_c(t)$, we aim to generate a landing trajectory $p(t)$ that is both smooth and dynamically feasible. To avoid potential interference from the cable, the trajectory optimization problem is formulated as:
\begin{subequations}\label{eq.16}
\begin{align}
&   {\min _{p(t), T} \mathcal{J}_o=\int_0^T\left\|\mathbf{p}^{(s)}(t)\right\|^2 \mathrm{~d} t+\rho T}, \label{eq.16a} \\
&   {\mathrm{~s} . \mathrm{t.} T>0},\label{eq.16b}\\
&   {\mathbf{p}^{[-1]}(0)=\mathbf{p}_o},\label{eq.16c}\\
&   {\mathbf{p}(T)=\mathbf{p}_g},\label{eq.16d}\\
&   {\mathbf{p}^{(1)}(T)=\mathbf{p}_c^{(1)}(T)},\label{eq.16e}\\
&   {\left\|\mathbf{p}^{(1)}(t)\right\| \leq v_{\max }, \forall t \in[0, T]},\label{eq.16f}\\
&   {\tau_{\min } \leq\|\boldsymbol{\tau}(t)\| \leq \tau_{\max }, \forall t \in[0, T]},\label{eq.16g}\\
&   {l_{\min }[p(t)] \leq l_{\text {now }}(t) \leq l_{\max }[p(t)], \forall t \in[0, T]}.\label{eq.16h}
\end{align}
\end{subequations}

Here, $T$ denotes the total trajectory duration, and $\mathbf{p}_o$ represents the initial state of the end droid, including position, velocity, acceleration, and jerk, $\mathbf{p}_c$ denotes the target point. In this formulation,  (\ref{eq.16c}) imposes the initial state constraint, while  (\ref{eq.16d}) and (\ref{eq.16e}) enforce the terminal position and velocity constraints. These ensure that the UAV reaches the target safely and smoothly.  (\ref{eq.16f}) and (\ref{eq.16g}) represent the velocity and net thrust constraints, and  (\ref{eq.16h}) corresponds to the cable length constraint.

To solve the above constrained optimization problem, the terminal constraints are relaxed and the continuous-time constraints are penalized. This reformulates the original problem as an unconstrained optimization, expressed as:
\begin{equation}
\min _{p(t), T} \mathcal{J}=\mathcal{J}_o+\sum w_i \mathcal{J}_i,
\end{equation}
where $\mathcal{J}_i$ denotes the cost function corresponding to each constraint, $i = \{v, \tau, \mathcal{L}\}$, including penalties on velocity, thrust, and cable length. The $w_i$ are weight coefficients that balance the influence of each term. By computing the gradients of the unconstrained cost function with respect to the polynomial coefficients and trajectory duration, the trajectory can be iteratively optimized using a quasi-Newton optimizer\cite{ref26}.

The smoothness cost $\mathcal{J}_s$ is defined as the integral of the squared $s$-th derivative of the trajectory, capturing the control effort associated with higher-order motion:
\begin{equation}
\mathcal{J}_s = \int_{t_0}^{t_M} \left\| p^{(s)}(t) \right\|_2^2 \, dt,
\end{equation}

Here, $t \in [t_0, t_M]$ denotes the trajectory time domain.

To encourage time efficiency, a time penalty term is introduced:
\begin{equation}
\mathcal{J}_t = \mathrm{sum}(\mathbf{T}) = t_M - t_0.
\end{equation}

In addition to smoothness and time, dynamic feasibility must be ensured through velocity, acceleration and jerk constraints. To allow soft enforcement of these constraints, we define the following penalty functions based on discrete-time sampling:
\begin{align}
\mathcal{J}_{d,v} &= \sum_{i=0}^{\kappa} \max\left\{ \left\| \dot{p}(t_i) \right\|^2 - v_m^2, \, 0 \right\}^3 ,\\
\mathcal{J}_{d,a} &= \sum_{i=0}^{\kappa} \max\left\{ \left\| \ddot{p}(t_i) \right\|^2 - a_m^2, \, 0 \right\}^3 ,\\
\mathcal{J}_{d,j} &= \sum_{i=0}^{\kappa} \max\left\{ \left\| \dddot{p}(t_i) \right\|^2 - j_m^2, \, 0 \right\}^3 ,\\
\mathcal{J}_d &= \mathcal{J}_{d,v} + \mathcal{J}_{d,a} + \mathcal{J}_{d,j}.
\end{align}
where $v_m$, $a_m$, and $j_m$ are the maximum allowable velocity, acceleration and jerk, respectively. Each sample time is defined as:
\begin{equation}
t_i = t_0 + \frac{t_M - t_0}{\kappa} \cdot i,
\end{equation}

Obstacle avoidance is performed in an unstructured environment with planar obstacles modeled as:
\begin{equation}
(\mathbf{x} - \mathbf{s})^\top \mathbf{v} = 0,
\end{equation}
where $\mathbf{x} \in \mathbb{R}^3$ and $\mathbf{s} \in \mathbb{R}^3$ is a point on the obstacle surface. The signed distance to the obstacle is given by:
\begin{equation}
d_o = (\mathbf{p} - \mathbf{s})^\top \mathbf{v},
\end{equation}
where $\mathbf{v}$ is the unit normal pointing toward the free space. If $d_o < \mathcal{C}_o$ for some positive safety margin $\mathcal{C}_o > 0$, a soft obstacle avoidance penalty is introduced:
\begin{equation}
\mathcal{J}_o = \sum_{i=0}^{\kappa} \max\left\{ \mathcal{C}_o - d_o\left(p(t_i)\right), \, 0 \right\}^3 .
\end{equation}

Cable length constraints are enforced to ensure that the UAV is neither excessively pulled nor slackened. The cable penalty is defined as:
\begin{equation}
\mathcal{J}_{\mathcal{L}} = 
\begin{cases}
\sum_{i=0}^{\kappa} \max\left\{ L_{\min}^2(t) - L^2(t), 0 \right\}^3, & L < L_{\min}, \\
\sum_{i=0}^{\kappa} \max\left\{ L^2(t) - L_{\max}^2(t), 0 \right\}^3, & L > L_{\max},
\end{cases}
\end{equation}
here, $L_{\min}(t)$ is the minimum allowable cable length, computed as the horizontal-plane distance from the UAV to the cable attachment point:
\begin{equation}
L_{\min}^2(t) = \left\| \mathbf{p}(t) - \mathbf{p}_0 \right\|_{xz}^2,
\end{equation}
$L_{\max}(t)$ is the maximum allowed cable length, derived from the catenary model with sag, and $L(t)$ is the instantaneous cable length during flight.

\section{Experiments and Results}

\subsection{Simulation experiment}
We developed a simulation environment using the Gazebo\cite{krusi2017driving} platform to evaluate the effectiveness and practicality of the proposed system. To simplify the analysis, we made the assumption that the payload drone is initially positioned near the target area and is stably hovering. From this position, the end droid move towards the target point.

\begin{figure}[!t]
\centering
\includegraphics[width=3.4in]{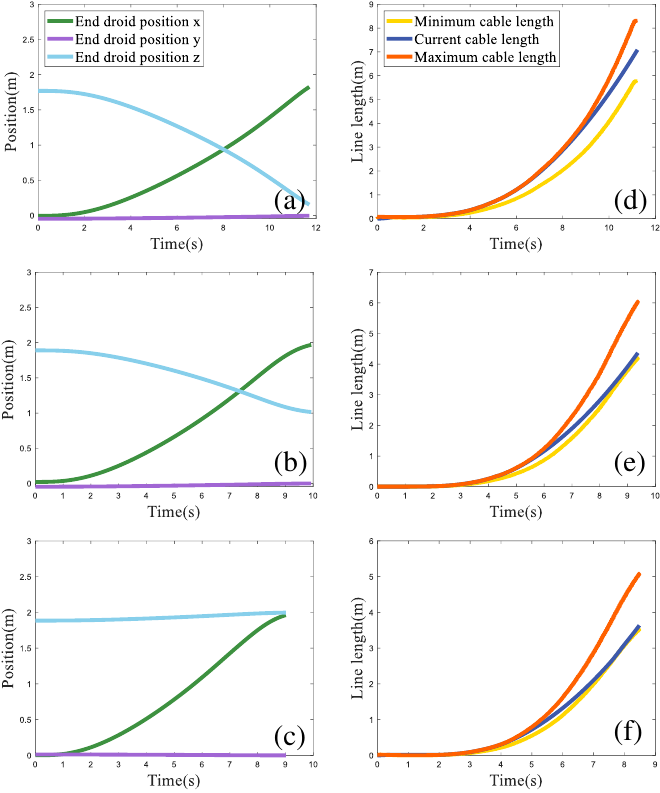}
\caption{Trajectory planning results and corresponding cable length profiles of the end droid under three target point configurations. (a)-(c) The planned position trajectories of the end droid along x, y, and z axes under three target point settings: (2, 0, 0)~m, (2, 0, 1)~m, and (2, 0, 2)~m, respectively. The initial position is fixed at (0, 0, 0)~m. (d)-(f) The corresponding cable length variations, including the minimum, maximum, and current cable lengths. In all cases, the planned motion remains within the allowable cable length bounds, validating the feasibility of the proposed planner under tether constraints.}
\label{fig_8}
\end{figure}

To assess the feasibility of the proposed trajectory planning method under varying target altitudes, we conducted three sets of simulation experiments. In each experiment, the initial position of end droid is set at the origin (0, 0, 0)~m, with the target points placed at (2, 0, 0)~m, (2, 0, 1)~m, and (2, 0, 2)~m. The corresponding planned trajectories are illustrated in Fig.~\ref{fig_8}(a)-(c), with the cable length variations shown in Fig.~\ref{fig_8}(d)-(f). The physical parameters of the cable, such as sag, unit length weight, and payout speed, are provided in Table \ref{tab_1}.

\begin{table}[ht]
\centering
\caption{Physical Parameters of the Cable}
\label{tab_1}
\begin{tabular}{lc}
\toprule
\textbf{Parameter} & \textbf{Value} \\
\midrule
Sag ($d$) & 0.1~m \\
Unit weight (per meter) & 0.14~g/m \\
Pay-out speed & 0.2~m/s \\
\bottomrule
\end{tabular}
\end{table}

\begin{figure}[!t]
\centering
\includegraphics[width=3.4in]{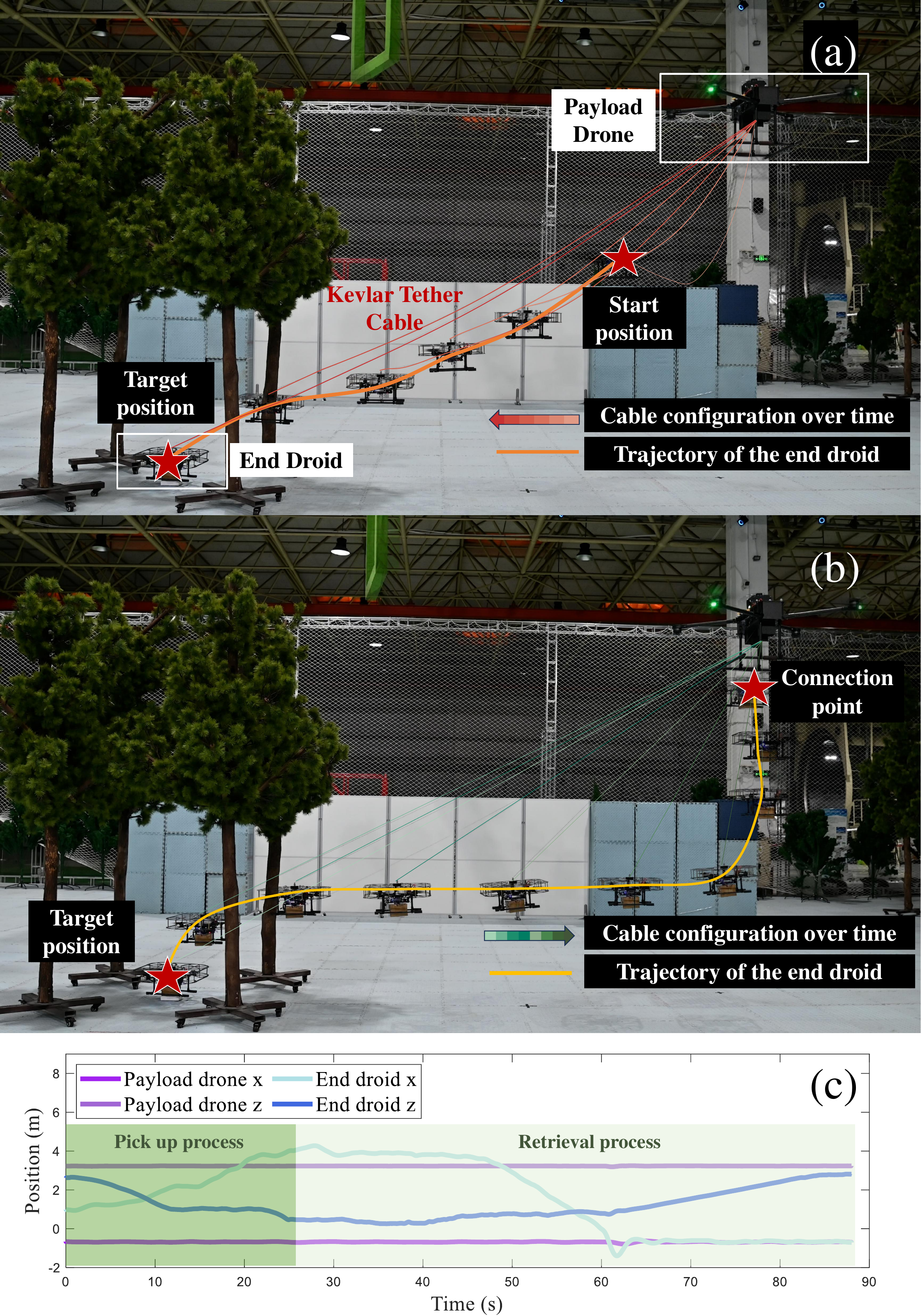}
\caption{Overview of the system architecture and experimental setup. (a) Pick-up process. (b) Retrieval process after pick-up. (c) $x$–$z$ position trajectories of the payload drone and the end droid.}
\label{fig_9}
\end{figure}

The simulation results demonstrate that, regardless of the target point’s altitude, the cable length required by the end droid during its flight remains consistently within the pre-established maximum and minimum limits. This behavior indicates that the cable system, under the given constraints, does not significantly hinder the mobility of the end droid. These findings underscore the system’s strong adaptability in trajectory planning and its capability to execute tasks effectively, even with cable length constraints. This confirms that the proposed system is capable of achieving safe, efficient motion planning in practical applications.

\subsection{Real-World experiment}
To validate the effectiveness of the proposed method in real-world scenarios, we conducted flight experiments using the custom aerial platform described in Section~III. As shown in Fig.~\ref{fig_9}, the system includes a payload drone and a rappelling end droid operating in a motion-capture-based indoor environment. The experiments evaluate the system's ability to perform constrained trajectory planning and robust descent under different target conditions. The physical parameters of the cable, such as sag, unit length weight, and payout speed, are consistent with those used in the simulation and are listed in Table~\ref{tab_1}.

In real-world experiments, densely branched forest environments impose severe constraints on large UAVs, making it difficult for them to land or directly approach target areas. To simulate a cluttered natural environment, we constructed a simplified ``forest'' in an indoor space using four trees arranged in a rectangular layout. A static target object was placed at the center of the area enclosed by the four trees, creating significant visual occlusion and physical constraints. Due to the high density of obstacles, the payload drone could not access the target location directly from above. Instead, the end droid was required to guide the tether cable through the narrow space to complete the pick-up task. This experimental setup is designed to verify the cooperative aerial system’s capability for navigating constrained environments, the accuracy of the cable modeling, and the overall task execution performance.

At the beginning of the experiment, the end droid was stationary and suspended directly below the payload drone. After the experiment started, the payload drone transported the system to a location near the target area. The end droid then performed real-time trajectory planning based on its own state, the current cable length, and the payload drone’s position (provided by the motion capture system). It gradually descended and maneuvered through the obstacle area to reach the target point and complete the pick-up.

After successfully picking up the target, the tether cable was retracted via the winch, passively pulling the end droid back to the position beneath the payload drone. During the entire process, trajectory planning accounted for cable length constraints to ensure safe and feasible motion, enabling the end droid to fly safely while carrying the cable.

\begin{figure}[!t]
\centering
\includegraphics[width=2.0in]{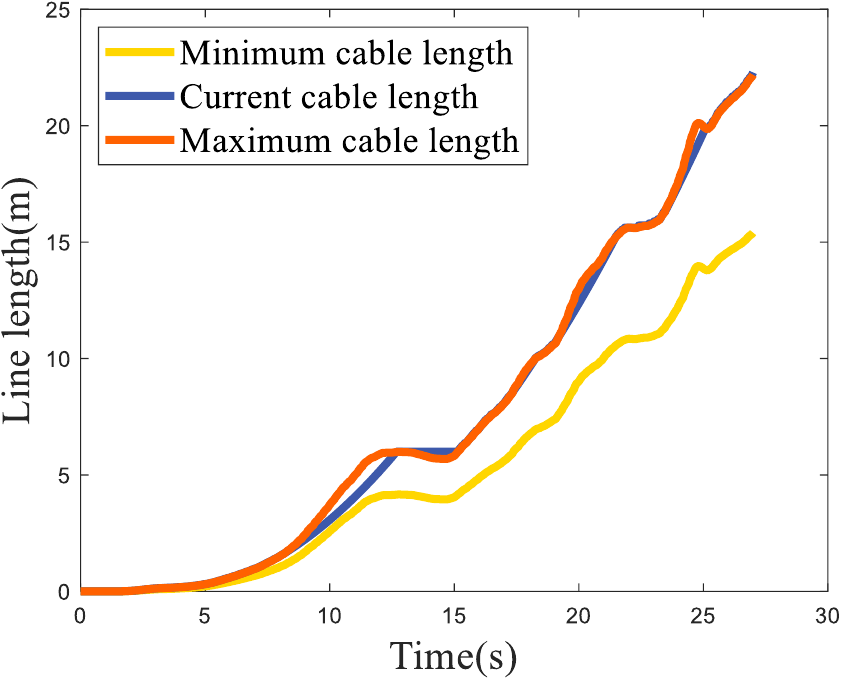}
\caption{Cable length constraint during the pick-up process.}
\label{fig_12}
\end{figure}

The experimental results are shown in Fig.~\ref{fig_9}, which illustrates the pick-up process, the post-grasp retrieval process, and the position trajectories of both drones in the $Oxz$ plane. The time interval from 0 to 27~s corresponds to the pick-up phase, and 27 to 88~s corresponds to the retrieval phase. Fig.~\ref{fig_12} presents the cable length constraint during the pick-up phase. The results indicate that the real-time cable length remained within the specified upper and lower bounds throughout the process, and the end droid successfully completed the task while safely handling the cable.

\section{Conclusion}
In this paper, we introduce a novel cooperative aerial system comprising a payload UAV and a dexterous rappelling end droid connected by a cable, designed to address the challenges of picking targets hidden behind obstacles and at varying distances. The end droid enables controlled descent to the target, establishes a secure connection with the payload, and allows the system to interact with objects in complex environments. To ensure safe and efficient operation, we employ a catenary model to establish maximum safe cable length constraints. A trajectory optimization method for the end droid's motion under cable constraints, validated through both simulation and real-world experiments. The system expands the operational workspace compared to traditional mechanisms, avoids the complexities of mid-air retrieval, and mitigates aerodynamic effects. Future research will focus on optimizing the fine operation of the end droid, incorporating multi-UAV collaboration and visual perception to expand the system's applications in dynamic target handling and collaborative transport tasks.

\bibliographystyle{IEEEtran}
\bibliography{IEEEexample}

\vfill

\end{document}